\documentclass[conference]{IEEEtran}
\IEEEoverridecommandlockouts
\usepackage{cite}
\usepackage{amsmath,amssymb,amsfonts}
\usepackage{algorithm}
\usepackage{algorithmic}
\usepackage{graphicx}
\usepackage{textcomp}
\usepackage{xcolor}

\usepackage{amsmath}
\usepackage{amsfonts}
\usepackage{multirow}
\usepackage{amssymb}
\usepackage{booktabs}
\usepackage{pifont}   % Required for \ding command
\usepackage{wasysym}  % Required for \Square and \CheckedBox commands

\usepackage{caption}
\renewcommand{\tablename}{Table} % 更改表格标题的“TABLE”部分
\renewcommand{\thetable}{\arabic{table}} % 使用阿拉伯数字代替罗马数字
\renewcommand{\thefigure}{\arabic{figure}}  % 使用阿拉伯数字代替罗马数字
\def\BibTeX{{\rm B\kern-.05em{\sc i\kern-.025em b}\kern-.08em
    T\kern-.1667em\lower.7ex\hbox{E}\kern-.125emX}}
\begin{document}

\title{EasySplat: View-Adaptive Learning makes 3D Gaussian Splatting Easy}

% \author{Anonymous ICME submission}
\author{
    Ao Gao \textsuperscript{1}, Luosong Guo \textsuperscript{2}, Tao Chen \textsuperscript{3}, Zhao Wang\textsuperscript{3}, Ying Tai \textsuperscript{1}, Jian Yang \textsuperscript{1}, Zhenyu Zhang \textsuperscript{1*} \\
    % 如果有更多作者，继续添加
    % 第三作者姓名 \textsuperscript{3} \\
    % 作者所属机构
    \textsuperscript{1} Nanjing University \\
    \textsuperscript{2} Nanjing University of Aeronautics and Astronautics \\
    \textsuperscript{3} China Mobile \\
    \small 2202521@mail.dhu.edu.cn, Luosongguo@nuaa.edu.cn, chentao@js.chinamobile.com, wangzh8@js.chinamobile.com, \\ yingtai@nju.edu.cn, csjyang@nankai.edu.cn, zhenyuzhang@nju.edu.cn \\
}

\maketitle

\begin{abstract}
3D Gaussian Splatting (3DGS) techniques have achieved satisfactory 3D scene representation. Despite their impressive performance, they confront challenges due to the limitation of structure-from-motion (SfM) methods on acquiring accurate scene initialization, or the inefficiency of densification strategy. In this paper, we introduce a novel framework EasySplat to achieve high-quality 3DGS modeling. Instead of using SfM for scene initialization, we employ a novel method to release the power of large-scale pointmap approaches. Specifically, we propose an efficient grouping strategy based on view similarity, and use robust pointmap priors to obtain high-quality point clouds and camera poses for 3D scene initialization. After obtaining a reliable scene structure, we propose a novel densification approach that adaptively splits Gaussian primitives based on the average shape of neighboring Gaussian ellipsoids, utilizing KNN scheme. In this way, the proposed method tackles the limitation on initialization and optimization, leading to an efficient and accurate 3DGS modeling. Extensive experiments demonstrate that EasySplat outperforms the current state-of-the-art (SOTA) in handling novel view synthesis. 
\end{abstract}


\section{Introduction}
\label{sec:intro}

Novel View Synthesis (NVS) is a challenging task in computer vision and computer graphics. Recently, neural rendering techniques have gained prominence due to their superior ability to achieve highly realistic renderings. Among these techniques, 3D Gaussian Splatting (3DGS) ~\cite{3dgs}, which employs an explicit point-cloud representation, has demonstrated state-of-the-art performance in both rendering quality and speed.

\begin{figure}[!ht]
  \centerline{\includegraphics[width=1.0\columnwidth]{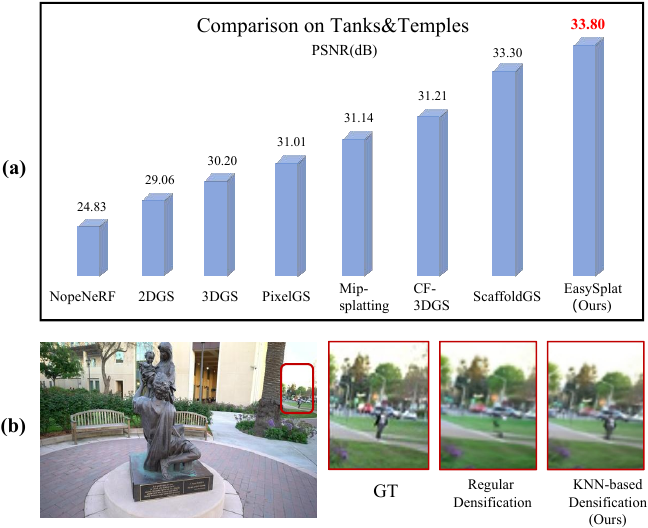}}
  \caption{\textbf{Comparison with existing methods.} (a) Compared with other SOTA methods, our method achieves the best performance in rendering quality. (b) In contrast to the regular densification used in the vanilla 3DGS, our KNN-based densification effectively grows points in areas where the initial point cloud is insufficient, leading to more accurate and detailed results.}
  \label{fig1}
\end{figure}

3DGS uses the Structure-from-Motion (SfM) method, COLMAP~\cite{schonberger2016colmap}, to extract camera poses and an initial sparse point cloud from hundreds of images and achieves real-time realistic rendering through a differentiable rasterizer. Despite the high-quality novel view synthesis performance, it often produces noisy Gaussians due to two major limitations. One reason is the use of SfM as an initialization method, which introduces noise into the point cloud due to its sensitivity to feature extraction errors and the difficulty in handling textureless scenarios~\cite{cf3dgs}, significantly degrading the final view synthesis and rendering quality. Recently, pointmap-based Multi-View Stereo (MVS) models DUSt3R\cite{wang2024dust3r} has shown excellent performance in dense 3D reconstruction. By adopting an end-to-end estimation process based on Transformer models, it can easily obtain pairwise pointmaps, which can be used to represent geometric relationships between two images. InstantSplat ~\cite{fan2024instantsplat} utilizes DUSt3R for initialization to learn 3DGS from sparse views. However, it relies on constructing a complete connectivity graph between views, limiting its application in dense-view scenes due to the significant cost of time and space. As a result, how to perform suitable scene initialization or estimate camera poses for 3DGS is still an opening problem. 

Besides the initialization, the training strategy in 3DGS is another reason for the sub-optimal performance. Since SfM techniques often fail to generate sufficient 3D points in textureless regions, 3DGS implements an Adaptive Density Control (ADC) algorithm to manage Gaussian primitives. This algorithm performs point densification and pruning regularly based on a view-average gradient magnitude threshold~\cite{yang2024lpm}. However, the less-constrained densification cannot effectively grow points in areas where the initial point cloud is sparse, finally degrading the rendering quality. To overcome these limitations, ScaffoldGS ~\cite{lu2024scaffoldgs} and OctreeGS ~\cite{ren2024octreegs} are proposed to dynamically generate neural Gaussians by introducing anchor-based structures. Moreover, Mip-Splatting ~\cite{yu2024mip_splat} introduces low-pass filtering to address high-frequency artifacts. 

In this paper, we propose EasySplat, a framework for achieving high-quality novel view synthesis. Specifically, we introduce an adaptive grouping strategy based on image similarity for the initialization of dense-view scenes. Subsequently, we employ the K-Nearest Neighbors (KNN) algorithm to identify the N closest Gaussians to each individual Gaussian. We then compute the average shape of these neighboring Gaussians. By comparing the discrepancies between the Gaussian shapes and this computed average shape, we determine whether a given Gaussian should be subdivided. As can be observed from Figure.~\ref{fig1}(b), our KNN-based densification effectively densifies Gaussians in regions with insufficient initial points. In this way, the novel 3DGS learning framework we proposed releases the power of the large-scale MVS models, enhancing the NVS efficiency as well as the performance.

In summary, our contributions are as follows:

\begin{itemize}
    \item We introduce EasySplat, a 3D Gaussian Splatting-based framework for NVS that outperforms the state-of-the-art in terms of NVS rendering quality and training speed.
    \item We propose a novel view-adaptive grouping strategy and leverage powerful pointmap priors to construct pairwise pointmap, thereby achieving precise initialization of point clouds and camera poses.
    \item We develop an adaptive densification strategy using KNN algorithm, which dynamically triggers densification in response to the shape discrepancies of adjacent ellipsoids for each Gaussian, thereby achieving robust novel view synthesis.
\end{itemize}

\begin{figure*}[ht]
\centering
\includegraphics [width=1.0\textwidth]{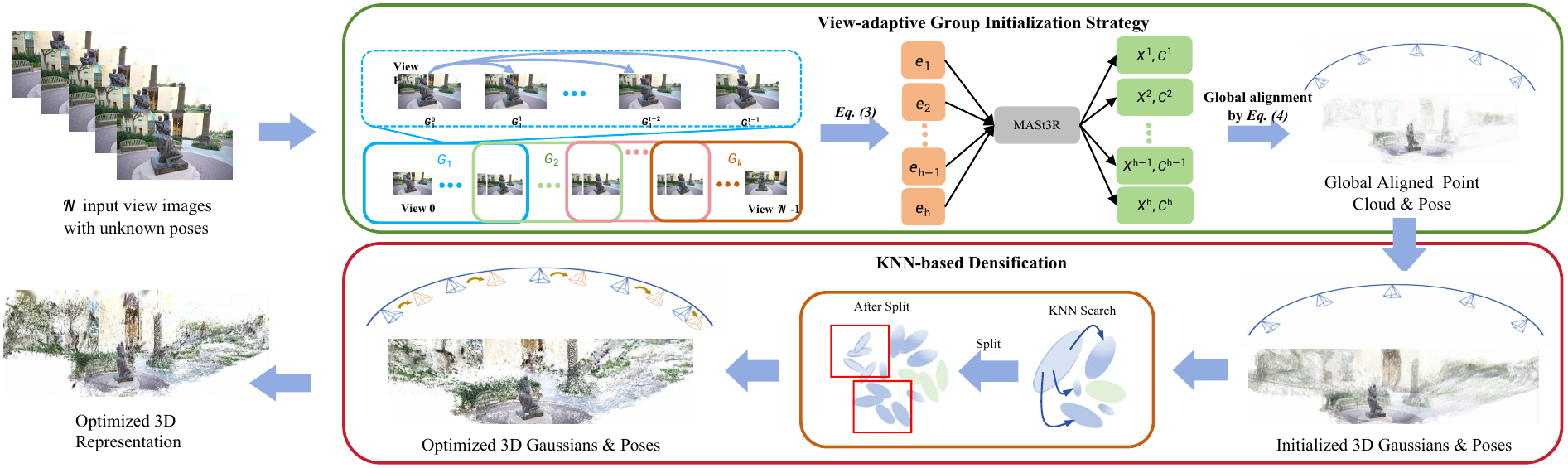} % Reduce the figure size so that it is slightly narrower than the column.
\caption{\textbf{Overview of proposed EasySplat.} Given $N$ images, we first construct image pairs based on view similarity to estimate paired point clouds, followed by global alignment to estimate camera poses and point clouds. During training, we use a KNN-based adaptive division to control the density of Gaussian distributions while optimizing camera poses.}
\label{fig2}
\end{figure*}

\section{Related work}

\noindent\textbf{Novel View Synthesis.}
Neural Radiance Fields (NeRF) ~\cite{mildenhall2021nerf} is a pioneering method in the field of novel view synthesis (NVS), utilizing a Multi-Layer Perceptron (MLP) for scene modeling and employing volumetric rendering ~\cite{drebin1988volumerendering} for high-quality rendering performance. Follow-up works improve upon the NeRF method by enhancing training speed ~\cite{ muller2022instantngp}, rendering speed ~\cite{yu2021plenoctrees} and  rendering quality ~\cite{barron2022mipnerf360}. However, these methods either sacrifice speed or render quality.

Recently, 3DGS utilizes anisotropic Gaussians ~\cite{zwicker2001ewa} as representation, has demonstrated significant performance in both speed and rendering quality~\cite{yan2024multiscalegs}. Based on this explicit representation, several variant methods have been proposed~\cite{niemeyer2024radsplat}. FSGS~\cite{zhu2023fsgs} and DNGaussian~\cite{li2024dngaussian} have been proposed to learn Gaussian parameters with a limited number of images. GaussianPro ~\cite{cheng2024gaussianpro} develop a progressive propagation strategy to guide the densification of the 3D Gaussians. Pixel-GS ~\cite{zhang2024pixelgs} proposes a gradient scaling technique to mitigate artifacts close to the camera. FreGS ~\cite{zhang2024fregs} achieves Gaussian densification through a coarse-to-fine frequency annealing method. Several methods ~\cite{lu2024scaffoldgs} introduce structured grid features to dynamically generate neural Gaussians. Additionally, some works have attempted to extract 3D surfaces from Gaussian Splatting ~\cite{chen2023neusg}. SuGaR~\cite{guedon2024sugar} employs planar Gaussians aligned with object surfaces. 2DGS~\cite{huang20242dgs} utilizes planar 2D Gaussians primitives as representation.

\noindent\textbf{Efficient Prior for Novel View Synthesis.} 
Although SfM provides effective initialization for NeRF and 3DGS, it requires dense image capture, and when the captured images lack sufficient overlap and rich textures, SfM may introduce cumulative errors or even fail ~\cite{fan2024instantsplat}. To reduce reliance on SfM initialization, some methods have begun to simultaneously optimize camera poses and NeRF training ~\cite{wang2021nerfmm}. Nope-NeRF ~\cite{bian2023nopenerf} obtain distortion-free depth priors from monocular depth estimation and optimize both intrinsic and extrinsic camera parameters while training NeRF. The latest COLMAP-free Gaussian-based method, CF-3DGS ~\cite{cf3dgs}, estimates point clouds based on depth information and compute the relative poses of adjacent frames for training. However, these approaches typically require a long training time. An alternative approach is to replace COLMAP with more efficient pointmap-based prior models~\cite{wang2024dust3r}. InstantSplat combines DUSt3R with 3D Gaussian Splatting, exploring its application in sparse-view scenarios, and has achieved promising performance. Currently, no method has yet explored the combination of pointmap-based priors and 3D Gaussian Splatting in dense-view setting.

\section{Method}
\subsection{Overview}
In this section, we first introduce the overall framework of EasySplat, which can generate accurate 3D reconstruction representation by $N$ unposed images. As Figure. \ref{fig2} shows, given $N$ images \(\mathbf{I}_i \in \mathbb{R}^{H \times W \times 3}\) with unknown poses, they will be firstly handled by the Group Initialization Strategy to obtain the globally aligned pose and the global point cloud. Subsequently, the global point cloud is utilized to initialize the 3D Gaussian ellipsoids. During the densification phase, we employ K-Nearest Neighbor (KNN) search the K nearest ellipsoids for each Gaussian, determining whether a Gaussian should split by comparing the shape differences. The implementation details of these two key strategies will be elaborated in the following sections.

\subsection{View-adaptive Group Initialization Strategy}
To achieve effective initialization and address the limitations of SfM methods, we introduce DUSt3R~\cite{wang2024dust3r}, a powerful prior based on Transformer models, capable of generating point clouds from a pair of images. When confronted with dense viewpoint inputs, it is necessary to construct multiple pairwise image pairs, followed by the estimation of paired pointmaps. Finally, these pointmaps are globally aligned to obtain the globally aligned point cloud and camera poses. However, as shown in Table~\ref{tab:init_scheme_cmp}, we note that constructing a large number of pointmap pairs using the complete graph and swin methods consumes considerable memory, making it less suitable for dense views. Moreover, the camera poses derived from the pointmap pairs constructed by the oneref method are suboptimal, which may lead to a degradation in the performance of novel view synthesis.
 
\begin{table}[htbp]
\centering
\caption{\protect\textnormal{Ablation experiment on the Church scene of the Tanks\&Temples dataset, which consists of 400 images. \textbf{OOM denotes Out of Memory.} These results are reported on a single A6000 GPU.}}
\setlength{\tabcolsep}{4pt} % 设置列间距为4pt
\begin{tabular}{l cccccc}
\toprule
Scheme & PSNR↑ & ATE↓ & RPE\_t↓ & RPE\_r↓ & image pairs & GPU Mem \\
\midrule
complete & /     & /     & /     & /     & 159600 & OOM      \\
oneref   & 29.80 & \textbf{0.003} & 0.017 & 0.016 & 798    & 27510MB  \\
swin     & /     & /     & /     & /     & 2400   & OOM      \\
ours     & \textbf{30.22} & 0.005 & \textbf{0.015} & \textbf{0.013} & 1142   & 35118MB \\
\bottomrule
\end{tabular}
\label{tab:init_scheme_cmp}
\end{table}

To improve camera pose performance in dense-view scenes and avoid memory overhead, we propose an adaptive grouping strategy based on image similarity to construct image pairs. Given an input image sequence \( I = \{I_1, I_2, \dots, I_n\} \), we first compute the cosine similarity between each pair of adjacent images to quantify their similarity. The cosine similarity is defined as:

\begin{equation}
\text{sim}(I_i, I_{i+1}) = \frac{I_i \cdot I_{i+1}}{\|I_i\| \|I_{i+1}\|},
\end{equation}

where \( I_i \cdot I_{i+1} \) represents the dot product of images \( I_i \) and \( I_{i+1} \), and \( \|I_i\| \) and \( \|I_{i+1}\| \) are their respective norms. Next, we calculate the difference in similarity between adjacent images and construct a difference rate array \( \Delta \):

\begin{equation}
\Delta_i = \left| \text{sim}(I_i, I_{i+1}) - \text{sim}(I_{i+1}, I_{i+2}) \right| \quad i = 1, 2, \dots, n-2,
\end{equation}

where \( \Delta_i \) represents the difference in similarity between images \( I_i \) and \( I_{i+1} \).

Then, we select the \( k \) largest values from the difference rate array \( \Delta \). Finally, the image sequence \( I \) is divided into multiple subsequences \( I_1, I_2, \dots, I_k \), where each subsequence represents a group \(G\). Within each group \(G\), the view with an index of 0 is assigned as the reference view, and all other views within the group are matched to this reference view. This process generates a set of image pairs, denoted as Equation ~\eqref{eq:edge_equ}
\begin{equation}
e=(i,j|i=G_p^{0}, j=G_p^{q}, p \in [0, k], q \in [1, t)),
\label{eq:edge_equ}
\end{equation}

where $p$ is the group number, $G$ is the group set, $q$ is the non-reference number, $i$ and $j$ are pointmap number.

After pairing, all the image pairs are input to the DUSt3R's pretrained model to obtain the pairwise pointmaps \( X_{n,n} \), \( X_{m,n} \) and their associated confidence maps \( C_{n,n} \), \( C_{m,n} \) for each image pair $e \in E$. Then, a global optimization is performed as illustrated in Equation~\eqref{eq:global_align} to obtain the global point cloud and camera poses.

\begin{equation}
\begin{split}
\hat{J}^{*} = \arg \min_{\hat{J}, P, \sigma} & \sum_{e \in E} \sum_{v \in e} \sum_{i=1}^{HW} C^{v,e}_{i} \|\hat{J}^{v}_{i}-\sigma_e P_e X^{v,e}_{i}\|
\end{split}
\label{eq:global_align}
\end{equation}

By employing the view-adaptive grouping pairing strategy, we achieve more precise global point clouds and camera poses for subsequent 3DGS training.

\begin{table*}[htbp]
\centering
\caption{\textbf{Quantitative comparison of novel view synthesis results with previous SOTA methods on Tanks\&Temples}. The best results are highlighted in bold.}
\setlength{\tabcolsep}{3.8pt} % 设置列间距为4pt
\begin{tabular}{l ccc ccc ccc ccc ccc}
\toprule
\multirow{2}{*}{Scene} & \multicolumn{3}{c}{3DGS} & \multicolumn{3}{c}{CF-3DGS} & \multicolumn{3}{c}{Mip-Splatting} & \multicolumn{3}{c}{ScaffoldGS} & \multicolumn{3}{c}{EasySplat (Ours)} \\
\cmidrule(lr){2-4} \cmidrule(lr){5-7} \cmidrule(lr){8-10} \cmidrule(lr){11-13} \cmidrule(lr){14-16}
 & PSNR↑ & SSIM↑ & LPIPS↓ & PSNR↑ & SSIM↑ & LPIPS↓ & PSNR↑ & SSIM↑ & LPIPS↓ & PSNR↑ & SSIM↑ & LPIPS↓ & PSNR↑ & SSIM↑ & LPIPS↓ \\
\midrule
Church    & 29.93 & 0.93 & 0.09 & 30.54 & 0.93 & 0.09 & 30.49 & 0.94 & 0.08 & 31.95 & 0.95 & 0.08 & 30.22 & 0.94 & 0.08 \\
Barn      & 31.08 & 0.95 & 0.07 & 29.38 & 0.86 & 0.12 & 34.23 & 0.96 & 0.05 & 34.91 & 0.96 & 0.06 & 33.17 & 0.94 & 0.06 \\
Museum    & 34.47 & 0.96 & 0.05 & 29.45 & 0.91 & 0.10 & 35.16 & 0.97 & 0.04 & 35.04 & 0.97 & 0.04 & 35.62 & 0.97 & 0.04 \\
Family    & 27.93 & 0.92 & 0.11 & 33.47 & 0.96 & 0.05 & 30.20 & 0.94 & 0.08 & 31.62 & 0.95 & 0.06 & 35.23 & 0.97 & 0.03 \\
Horse     & 20.91 & 0.77 & 0.23 & 34.11 & 0.96 & 0.05 & 20.31 & 0.76 & 0.23 & 30.46 & 0.95 & 0.06 & 34.02 & 0.97 & 0.04 \\
Ballroom  & 34.48 & 0.96 & 0.04 & 32.47 & 0.96 & 0.07 & 35.11 & 0.97 & 0.03 & 35.36 & 0.98 & 0.03 & 36.62 & 0.98 & 0.02 \\
Francis   & 32.64 & 0.92 & 0.15 & 32.80 & 0.92 & 0.14 & 33.58 & 0.93 & 0.12 & 34.66 & 0.95 & 0.10 & 35.48 & 0.94 & 0.10 \\
Ignatius  & 30.20 & 0.93 & 0.08 & 27.46 & 0.90 & 0.09 & 30.03 & 0.93 & 0.07 & 32.36 & 0.95 & 0.06 & 30.04 & 0.91 & 0.08 \\
\midrule
Mean      & 30.205 & 0.918 & 0.103 & 31.210 & 0.925 & 0.089 & 31.139 & 0.925 & 0.088 & 33.295 &\textbf{ 0.956} & 0.064 & \textbf{33.800} & 0.953 & \textbf{0.056} \\
\bottomrule
\end{tabular}
\label{table:quantitative_comparison}
\end{table*}

\begin{table*}[h]
\centering
\caption{\textbf{Novel view synthesis and pose accuracy results on CO3DV2}. All results are evaluated using the same evaluation protocol. As for pose accuracy results, we use the camera poses provided by CO3DV2 as the “ground truth”. The best results are highlighted in bold.}
\setlength{\tabcolsep}{5.8pt} % 设置列间距为4pt
\begin{tabular}{l ccc ccc ccc ccc}
\toprule
\multirow{2}{*}{Scene} & \multicolumn{6}{c}{CF-3DGS} & \multicolumn{6}{c}{EasySplat(Ours)} \\
\cmidrule(lr){2-7} \cmidrule(lr){8-13}
 & PSNR↑ & SSIM↑ & LPIPS↓ & RPE\_t↓ & RPE\_r↓  & ATE↓  & PSNR↑ & SSIM↑ & LPIPS↓ & RPE\_t↓ & RPE\_r↓  & ATE↓  \\
\midrule

197\_21206\_41908 & 18.07  & 0.91  & 0.32  & \textbf{0.257}  & 1.592  & 0.065  & \textbf{30.35}  & \textbf{0.96}  & \textbf{0.23}  & 0.592  & \textbf{1.116}  & \textbf{0.015}  \\
219\_23121\_48537 & 25.09  & 0.80  & 0.40  & \textbf{0.112}  & \textbf{0.605}  & 0.027  & \textbf{31.73}  & \textbf{0.90}  & \textbf{0.20}  & 0.427  & 1.010  & \textbf{0.009}  \\ 
378\_43990\_87662 & 19.37  & 0.79  & 0.44  & \textbf{0.224}  &\textbf{ 0.867}  & 0.043  & \textbf{27.69}  & \textbf{0.90}  & \textbf{0.31}  & 1.107  & 1.719  & \textbf{0.023}  \\
437\_62536\_123478 & 16.44  & 0.67  & 0.46  & \textbf{0.206}  & 1.251  & 0.052  & \textbf{30.45}  & \textbf{0.91}   & \textbf{0.15}  & 0.434  & \textbf{0.902}  & \textbf{0.018}  \\

\bottomrule
\end{tabular}
\label{tab:co3dv2_comparison}
\end{table*}

\begin{figure}[!t]
\centering
\includegraphics[width=0.5\textwidth]{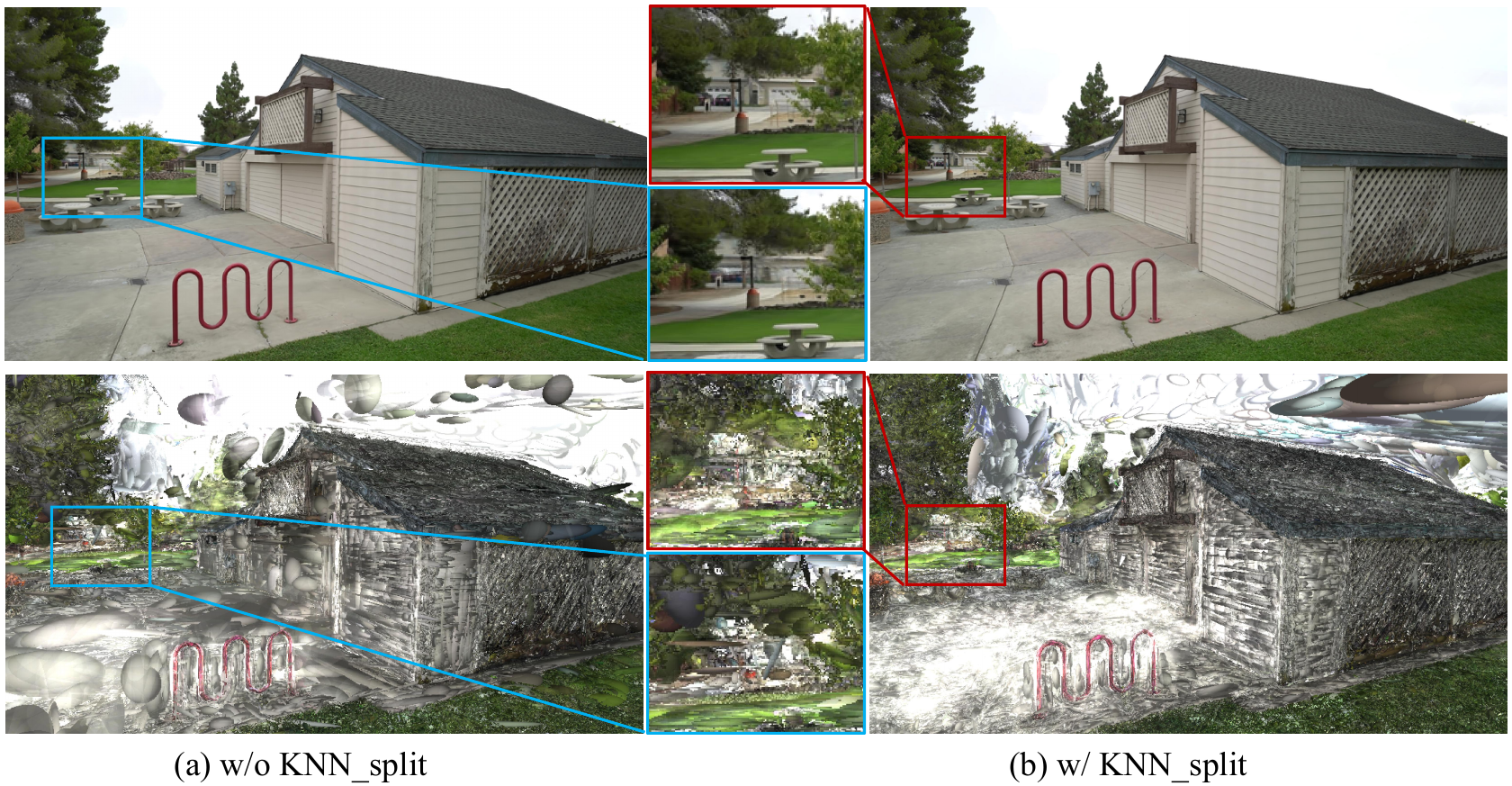} % Reduce the figure size so that it is slightly narrower than the column.
\caption{\textbf{KNN-based Densification}. After the KNN-based splitting, the large Gaussians are decomposed into smaller Gaussians, leading to significant improvements on smaller targets, such as the car depicted in the figure.}
\label{fig4}
\end{figure}

\subsection{KNN-based Densification}
Each 3D Gaussian primitive \( G_i(x) \) is composed of a mean vector \( \mu_{3d_i} \) and a full 3D covariance matrix \( \Sigma_{3d_i} \). The Gaussian primitive can be written as:
\begin{equation}
G_i(x) = e^{-\frac{1}{2} (x - \mu_{3d_i})^T \Sigma_{3d_i}^{-1} (x - \mu_{3d_i})}
\label{eq:gauss}
\end{equation}

Subsequently, during training, the average view-space positional gradient for each Gaussian primitive  \( G_i(x) \) is computed every 100 iterations. If the gradient exceeds the gradient threshold \( \tau_p \) and the shape exceeds the scale threshold \( \tau_S \), the Gaussian \( G_i(x) \) will undergo a split:

\begin{equation}
\nabla_{\mu_i} L > \tau_p \quad \text{and} \quad \Sigma_{3d_i} > \tau_S
\label{eq:split}
\end{equation}

Although ADC allows the split Gaussians to cover most of the scene, some large Gaussians tend to resist splitting. However, our observations in practical scenarios reveal that Gaussian distributions are uneven, with larger Gaussians often appearing in the proximity of smaller ones. Inspired by this phenomenon, we propose an adaptive Gaussian ellipsoid splitting strategy based on the K-Nearest Neighbors (KNN) algorithm. Specifically, for each Gaussian ellipsoid \( G_i \), We use the KNN algorithm to identify the \( n \) closest neighboring Gaussians \( G = \{ G_1, G_2, \dots, G_n \} \), and then compute the mean shape \( \bar{\Sigma}_{3d} \) of these neighbors.

\begin{equation}
\bar{\Sigma}_{3d} = \frac{1}{n} \sum_{j=1}^{n} \Sigma_{3d_j}
\label{eq:split}
\end{equation}

If \( \Sigma_{3d_i} > \bar{\Sigma}_{3d} \), the Gaussian \( G_i \) is classified as a large Gaussian and needs to be split. As shown in Figure~\ref{fig4}, through this splitting method, the large Gaussian is effectively divided.

\section{Experiments}

\subsection{Experimental Setup}
\noindent\textbf{Dataset} We conduct experiments on two real-world datasets: Tanks\&Temples~\cite{knapitsch2017tanks_temples}, and CO3DV2~\cite{reizenstein2021co3dv2}. \textbf{Tanks\&Temples:} We refer to CF-3DGS ~\cite{cf3dgs} and use 8 scenes to evaluate pose accuracy and novel view synthesis quality. For each scene, 7/8 of the images in each sequence are used for training, and the remaining 1/8 are used for testing the quality of novel view synthesis. Camera poses are estimated and evaluated after performing Umeyama alignment ~\cite{umeyama1991least} on all training samples. \textbf{CO3DV2:} CO3DV2 captures images by moving in a full circle around the target, with large and complex camera movements, making it more challenging to recover camera poses. We randomly choose four scenes and follow the same protocol as Tanks\&Temples to split the training/test set.

\begin{figure*}[!t]
\centering
\includegraphics[width=1.0\textwidth]{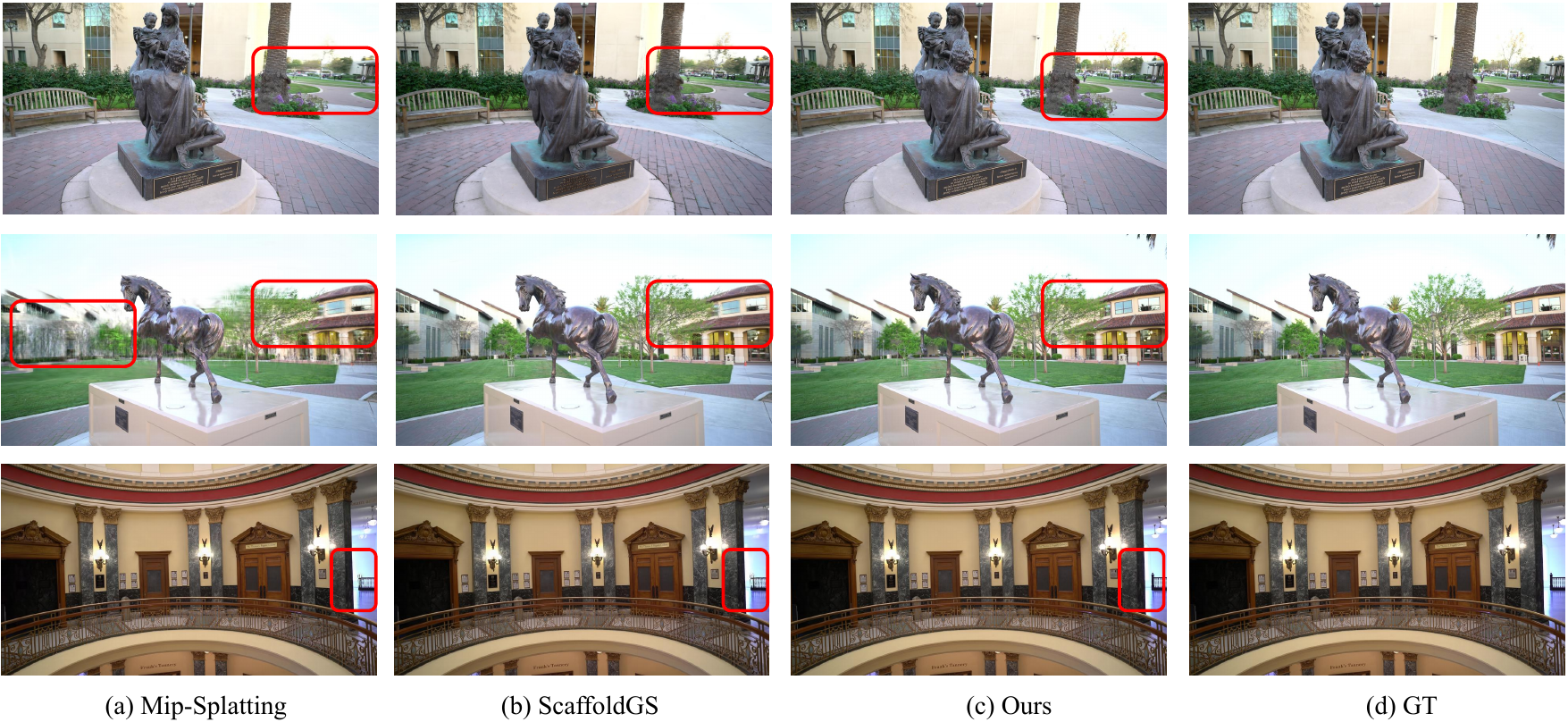} % Reduce the figure size so that it 
\caption{\textbf{Qualitative comparison for novel view synthesis on Tanks\&Temples.}  Our approach produces much more high-quality and detailed images than the baselines.
}
\label{fig3:visual_comparison}
\end{figure*}

\noindent\textbf{Metrics} We evaluate our approach on two primary tasks: novel view synthesis and camera pose estimation. For novel view synthesis, we follow previous methods ~\cite{cf3dgs} and use standard evaluation metrics including the Peak Signal-to-Noise Ratio (PSNR), the Structural Similarity Index Measure (SSIM) ~\cite{ssim}, and the Learned Perceptual Image Patch Similarity (LPIPS) ~\cite{lpips}. For camera pose estimation, we report standard evaluation metrics from visual odometry ~\cite{odometry}, which include the Absolute Trajectory Error (ATE), Relative Rotation Error (RPE$_r$), and Relative Translation Error (RPE$_t$).

\begin{figure}[!t]
\centering
\includegraphics[width=0.48\textwidth]{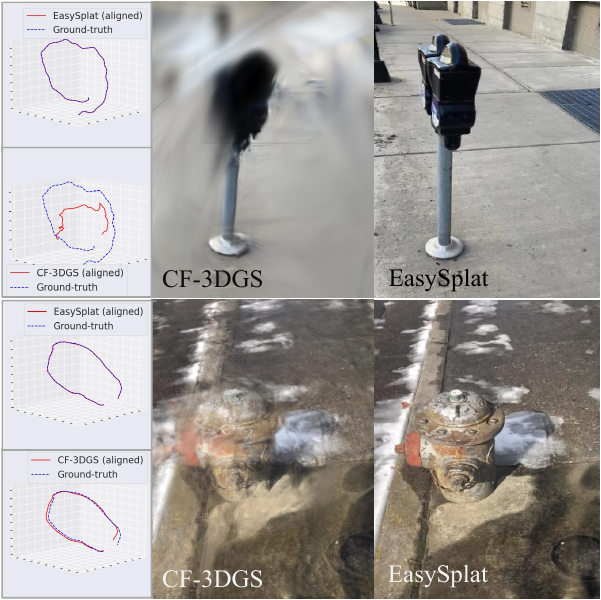} % Reduce the figure size so that it is slightly narrower than the column.
\caption{Qualitative comparison for novel view synthesis and camera pose estimation on CO3DV2.
}
\label{fig:co3dv2_comparison}
\end{figure}

\noindent\textbf{Implement Details} Our implementation utilizes the PyTorch framework. In constructing the pointmap groups, we set $k = 2$. The Gaussian neighborhood parameter $n$ is set to 64. For the pointmap prior model configuration, we employ the DUSt3R model ~\cite{leroy2024mast3r}, trained at a resolution of 512, using a ViT Large encoder and a ViT Base decoder. To ensure a fair comparison, all experiments are conducted on a single A6000 GPU. 

\subsection{Experimental Result}
\noindent\textbf{Quantitative and Qualitative Results} We conduct both qualitative and quantitative evaluations on the Tanks\&Temples dataset, comparing our method with current state-of-the-art (SOTA) methods, including 3DGS, CF-3DGS, Mip-Splatting, and ScaffoldGS. As shown in Table~\ref{table:quantitative_comparison}, our method achieves the best performance across all metrics. We also perform a qualitative evaluation, as illustrated in Figure~\ref{fig3:visual_comparison}. While the leading SOTA methods, ScaffoldGS and Mip-Splatting, demonstrate excellent novel view synthesis performance, they still exhibit blurriness and artifacts when changing views due to limitations imposed by COLMAP initialization. In contrast, our method, which incorporates more accurate initialization and view-adaptive learning strategies, delivers more sharp and clear visual results, demonstrating superior performance.

\begin{table}[ht]
\centering
\caption{Ablation study.}
\setlength{\tabcolsep}{6pt} % 设置列间距为4pt
\begin{tabular}{lccc}
\toprule
 scheme           & PSNR$\uparrow$ & SSIM$\uparrow$ & LPIPS$\downarrow$   \\
\midrule
 w/o View-adaptive Group Initialization   & 33.721          & 0.950           & 0.056              \\
 w/o KNN-based Densification   & 33.436          & 0.950           & 0.061     \\
 Full model      & \textbf{33.800}  & \textbf{0.953}   & \textbf{0.056}    \\
\bottomrule
\end{tabular}
\label{table:ablation}
\end{table}

\noindent\textbf{Results on Scenes with Large Camera Motions} To further evaluate EasySplat's performance in camera pose estimation, we present results on the CO3DV2 dataset, which comprises long videos with more complex camera movements. We select the existing optimal non-COLMAP initialization approach, CF-3DGS, as the comparison method. As shown in Table~\ref{tab:co3dv2_comparison} and Figure~\ref{fig:co3dv2_comparison}, our method significantly outperforms CF-3DGS in both novel view synthesis and camera pose estimation under the complex camera trajectories. 

\noindent\textbf{Ablation Studies} We conduct an ablation study focusing on the \textit{View-adaptive Group Initialization Strategy} and the \textit{KNN-based Densification}. When the group initialization is removed, we resort to an initialization method based on the oneref approach. As shown in Table~\ref{table:ablation}, there is a notable decline in metrics when both group initialization and KNN-based densification are omitted. Furthermore, as indicated in Table~\ref{tab:init_scheme_cmp}, the group initialization exhibits more accurate camera poses compared to the oneref approach, which in turn demonstrates superior performance in the NVS task.

\section{Conclusion \& Future Work}
In this paper, we propose EasySplat, a robust and efficient 3DGS-based framework for novel view synthesis (NVS). To address the issue of inaccurate sparse point cloud initialization caused by SfM in vanilla 3DGS, EasySplat utilizes an effective pointmap-based prior model for initialization. To release the power of pointmaps in dense-view scenarios, a group-based initialization strategy is proposed. Furthermore, to enhance the performance of NVS, we propose a densification scheme based on KNN algorithm. Extensive experiments demonstrate that EasySplat achieves the SOTA performance. In the future, this work could be extended to support both sparse and dense view settings, establishing a generalized 3DGS paradigm.

\bibliographystyle{IEEEbib}
\bibliography{icme2025references}

\begin{thebibliography}{10}

\bibitem{3dgs}
Bernhard M{\"u}ller, Georgios Kerbl, Thomas Kopanas, George Leimk¨uhler, and Drettakis,
\newblock ``3d gaussian splatting for real-time radiance field rendering,''
\newblock {\em ACM Transactions on Graphics (ToG)}, vol. 42, no. 4, pp. 1--14, 2023.

\bibitem{schonberger2016colmap}
Johannes~L Schonberger and Jan-Michael Frahm,
\newblock ``Structure-from-motion revisited,''
\newblock in {\em Proceedings of the IEEE conference on computer vision and pattern recognition}, 2016, pp. 4104--4113.

\bibitem{cf3dgs}
Yang, Sifei Fu, Amey Liu, Jan Kulkarni, Alexei~A Kautz, Xiaolong Efros, and Wang,
\newblock ``Colmap-free 3d gaussian splatting,''
\newblock {\em arXiv preprint arXiv:2312.07504}, 2023.

\bibitem{wang2024dust3r}
Shuzhe Wang, Vincent Leroy, Yohann Cabon, Boris Chidlovskii, and Jerome Revaud,
\newblock ``Dust3r: Geometric 3d vision made easy,''
\newblock in {\em Proceedings of the IEEE/CVF Conference on Computer Vision and Pattern Recognition}, 2024, pp. 20697--20709.

\bibitem{fan2024instantsplat}
Zhiwen Fan, Wenyan Cong, Kairun Wen, Kevin Wang, Jian Zhang, Xinghao Ding, Danfei Xu, Boris Ivanovic, Marco Pavone, Georgios Pavlakos, et~al.,
\newblock ``Instantsplat: Unbounded sparse-view pose-free gaussian splatting in 40 seconds,''
\newblock {\em arXiv preprint arXiv:2403.20309}, 2024.

\bibitem{yang2024lpm}
Haosen Yang, Chenhao Zhang, Wenqing Wang, Marco Volino, Adrian Hilton, Li~Zhang, and Xiatian Zhu,
\newblock ``Localized gaussian point management,''
\newblock {\em arXiv preprint arXiv:2406.04251}, 2024.

\bibitem{lu2024scaffoldgs}
Tao Lu, Mulin Yu, Linning Xu, Yuanbo Xiangli, Limin Wang, Dahua Lin, and Bo~Dai,
\newblock ``Scaffold-gs: Structured 3d gaussians for view-adaptive rendering,''
\newblock in {\em Proceedings of the IEEE/CVF Conference on Computer Vision and Pattern Recognition}, 2024, pp. 20654--20664.

\bibitem{ren2024octreegs}
Kerui Ren, Lihan Jiang, Tao Lu, Mulin Yu, Linning Xu, Zhangkai Ni, and Bo~Dai,
\newblock ``Octree-gs: Towards consistent real-time rendering with lod-structured 3d gaussians,''
\newblock {\em arXiv preprint arXiv:2403.17898}, 2024.

\bibitem{yu2024mip_splat}
Zehao Yu, Anpei Chen, Binbin Huang, Torsten Sattler, and Andreas Geiger,
\newblock ``Mip-splatting: Alias-free 3d gaussian splatting,''
\newblock in {\em Proceedings of the IEEE/CVF Conference on Computer Vision and Pattern Recognition}, 2024, pp. 19447--19456.

\bibitem{mildenhall2021nerf}
Ben Mildenhall, Pratul~P Srinivasan, Matthew Tancik, Jonathan~T Barron, Ravi Ramamoorthi, and Ren Ng,
\newblock ``Nerf: Representing scenes as neural radiance fields for view synthesis,''
\newblock {\em Communications of the ACM}, vol. 65, no. 1, pp. 99--106, 2021.

\bibitem{drebin1988volumerendering}
Robert~A Drebin, Loren Carpenter, and Pat Hanrahan,
\newblock ``Volume rendering,''
\newblock {\em ACM Siggraph Computer Graphics}, vol. 22, no. 4, pp. 65--74, 1988.

\bibitem{muller2022instantngp}
Thomas M{\"u}ller, Alex Evans, Christoph Schied, and Alexander Keller,
\newblock ``Instant neural graphics primitives with a multiresolution hash encoding,''
\newblock {\em ACM transactions on graphics (TOG)}, vol. 41, no. 4, pp. 1--15, 2022.

\bibitem{yu2021plenoctrees}
Alex Yu, Ruilong Li, Matthew Tancik, Hao Li, Ren Ng, and Angjoo Kanazawa,
\newblock ``Plenoctrees for real-time rendering of neural radiance fields,''
\newblock in {\em Proceedings of the IEEE/CVF International Conference on Computer Vision}, 2021, pp. 5752--5761.

\bibitem{barron2022mipnerf360}
Jonathan~T Barron, Ben Mildenhall, Dor Verbin, Pratul~P Srinivasan, and Peter Hedman,
\newblock ``Mip-nerf 360: Unbounded anti-aliased neural radiance fields,''
\newblock in {\em Proceedings of the IEEE/CVF conference on computer vision and pattern recognition}, 2022, pp. 5470--5479.

\bibitem{zwicker2001ewa}
Matthias Zwicker, Hanspeter Pfister, Jeroen Van~Baar, and Markus Gross,
\newblock ``Ewa volume splatting,''
\newblock in {\em Proceedings Visualization, 2001. VIS'01.} IEEE, 2001, pp. 29--538.

\bibitem{yan2024multiscalegs}
Zhiwen Yan, Weng~Fei Low, Yu~Chen, and Gim~Hee Lee,
\newblock ``Multi-scale 3d gaussian splatting for anti-aliased rendering,''
\newblock in {\em Proceedings of the IEEE/CVF Conference on Computer Vision and Pattern Recognition}, 2024, pp. 20923--20931.

\bibitem{niemeyer2024radsplat}
Michael Niemeyer, Fabian Manhardt, Marie-Julie Rakotosaona, Michael Oechsle, Daniel Duckworth, Rama Gosula, Keisuke Tateno, John Bates, Dominik Kaeser, and Federico Tombari,
\newblock ``Radsplat: Radiance field-informed gaussian splatting for robust real-time rendering with 900+ fps,''
\newblock {\em arXiv preprint arXiv:2403.13806}, 2024.

\bibitem{zhu2023fsgs}
Zehao Zhu, Zhiwen Fan, Yifan Jiang, and Zhangyang Wang,
\newblock ``Fsgs: Real-time few-shot view synthesis using gaussian splatting,''
\newblock {\em arXiv preprint arXiv:2312.00451}, 2023.

\bibitem{li2024dngaussian}
Jiahe Li, Jiawei Zhang, Xiao Bai, Jin Zheng, Xin Ning, Jun Zhou, and Lin Gu,
\newblock ``Dngaussian: Optimizing sparse-view 3d gaussian radiance fields with global-local depth normalization,''
\newblock in {\em Proceedings of the IEEE/CVF Conference on Computer Vision and Pattern Recognition}, 2024, pp. 20775--20785.

\bibitem{cheng2024gaussianpro}
Kai Cheng, Xiaoxiao Long, Kaizhi Yang, Yao Yao, Wei Yin, Yuexin Ma, Wenping Wang, and Xuejin Chen,
\newblock ``Gaussianpro: 3d gaussian splatting with progressive propagation,''
\newblock in {\em Forty-first International Conference on Machine Learning}, 2024.

\bibitem{zhang2024pixelgs}
Zheng Zhang, Wenbo Hu, Yixing Lao, Tong He, and Hengshuang Zhao,
\newblock ``Pixel-gs: Density control with pixel-aware gradient for 3d gaussian splatting,''
\newblock {\em arXiv preprint arXiv:2403.15530}, 2024.

\bibitem{zhang2024fregs}
Jiahui Zhang, Fangneng Zhan, Muyu Xu, Shijian Lu, and Eric Xing,
\newblock ``Fregs: 3d gaussian splatting with progressive frequency regularization,''
\newblock in {\em Proceedings of the IEEE/CVF Conference on Computer Vision and Pattern Recognition}, 2024, pp. 21424--21433.

\bibitem{chen2023neusg}
Hanlin Chen, Chen Li, and Gim~Hee Lee,
\newblock ``Neusg: Neural implicit surface reconstruction with 3d gaussian splatting guidance,''
\newblock {\em arXiv preprint arXiv:2312.00846}, 2023.

\bibitem{guedon2024sugar}
Antoine Gu{\'e}don and Vincent Lepetit,
\newblock ``Sugar: Surface-aligned gaussian splatting for efficient 3d mesh reconstruction and high-quality mesh rendering,''
\newblock in {\em Proceedings of the IEEE/CVF Conference on Computer Vision and Pattern Recognition}, 2024, pp. 5354--5363.

\bibitem{huang20242dgs}
Binbin Huang, Zehao Yu, Anpei Chen, Andreas Geiger, and Shenghua Gao,
\newblock ``2d gaussian splatting for geometrically accurate radiance fields,''
\newblock in {\em ACM SIGGRAPH 2024 Conference Papers}, 2024, pp. 1--11.

\bibitem{wang2021nerfmm}
Zirui Wang, Shangzhe Wu, Weidi Xie, Min Chen, and Victor~Adrian Prisacariu,
\newblock ``Nerf--: Neural radiance fields without known camera parameters,''
\newblock {\em arXiv preprint arXiv:2102.07064}, 2021.

\bibitem{bian2023nopenerf}
Wenjing Bian, Zirui Wang, Kejie Li, Jia-Wang Bian, and Victor~Adrian Prisacariu,
\newblock ``Nope-nerf: Optimising neural radiance field with no pose prior,''
\newblock in {\em Proceedings of the IEEE/CVF Conference on Computer Vision and Pattern Recognition}, 2023, pp. 4160--4169.

\bibitem{knapitsch2017tanks_temples}
Arno Knapitsch, Jaesik Park, Qian-Yi Zhou, and Vladlen Koltun,
\newblock ``Tanks and temples: Benchmarking large-scale scene reconstruction,''
\newblock {\em ACM Transactions on Graphics (ToG)}, vol. 36, no. 4, pp. 1--13, 2017.

\bibitem{reizenstein2021co3dv2}
Jeremy Reizenstein, Roman Shapovalov, Philipp Henzler, Luca Sbordone, Patrick Labatut, and David Novotny,
\newblock ``Common objects in 3d: Large-scale learning and evaluation of real-life 3d category reconstruction,''
\newblock in {\em Proceedings of the IEEE/CVF international conference on computer vision}, 2021, pp. 10901--10911.

\bibitem{umeyama1991least}
Shinji Umeyama,
\newblock ``Least-squares estimation of transformation parameters between two point patterns,''
\newblock {\em IEEE Transactions on Pattern Analysis \& Machine Intelligence}, vol. 13, no. 04, pp. 376--380, 1991.

\bibitem{ssim}
Zhou Wang, Alan~C Bovik, Hamid~R Sheikh, and Eero~P Simoncelli,
\newblock ``Image quality assessment: from error visibility to structural similarity,''
\newblock {\em IEEE transactions on image processing}, vol. 13, no. 4, pp. 600--612, 2004.

\bibitem{lpips}
Richard Zhang, Phillip Isola, Alexei~A Efros, Eli Shechtman, and Oliver Wang,
\newblock ``The unreasonable effectiveness of deep features as a perceptual metric,''
\newblock in {\em Proceedings of the IEEE conference on computer vision and pattern recognition}, 2018, pp. 586--595.

\bibitem{odometry}
Zichao Zhang and Davide Scaramuzza,
\newblock ``A tutorial on quantitative trajectory evaluation for visual (-inertial) odometry,''
\newblock in {\em 2018 IEEE/RSJ International Conference on Intelligent Robots and Systems (IROS)}. IEEE, 2018, pp. 7244--7251.

\bibitem{leroy2024mast3r}
Vincent Leroy, Yohann Cabon, and J{\'e}r{\^o}me Revaud,
\newblock ``Grounding image matching in 3d with mast3r,''
\newblock {\em arXiv preprint arXiv:2406.09756}, 2024.

\end{thebibliography}

\vspace{12pt}
\end{document}